# Clustering Tree-structured Data on Manifold

## Na Lu, Hongyu Miao

**Abstract**—Tree-structured data usually contain both topological and geometrical information, and are necessarily considered on manifold instead of Euclidean space for appropriate data parameterization and analysis. In this study, we propose a novel tree-structured data parameterization, called Topology-Attribute matrix (T-A matrix), so the data clustering task can be conducted on matrix manifold. We incorporate the structure constraints embedded in data into the negative matrix factorization method to determine meta-trees from the T-A matrix, and the signature vector of each single tree can then be extracted by meta-tree decomposition. The meta-tree space turns out to be a cone space, in which we explore the distance metric and implement the clustering algorithm based on the concepts like Fréchet mean. Finally, the T-A matrix based clustering (TAMBAC) framework is evaluated and compared using both simulated data and real retinal images to illustrate its efficiency and accuracy.

## 1 INTRODUCTION

TREE-STRUCTURED data are common and play an important role in our lives. Representative examples in biomedical research and clinical practice include blood vessel systems [1], human pulmonary airway structures from CT scans [2], and phylogenetic trees of biological species or molecules [3]. In particular, abundant angiographic images [4] are generated every day to facilitate the diagnosis and treatment of tens of diseases such as diabetic retinopathy, tumor, and cardiovascular disease [5]. Further examples include file directory trees in computer science [6], family tress in social behavioral science [7], leaf vasculatures in plant biology [8] and so on.

Quantitative analysis of tree-structured data presents a unique challenge primarily for three reasons. First, appropriate data parameterizations need to be explored to reflect the rich information embedded in tree-structured data. More specifically, tree-structured data usually contain both topological information (i.e., parent-child relationships) and certain attributes associated with each tree node or edge (i.e., branch length). Therefore, it is unlikely for any single quantitative measure to sufficiently represent a tree. For instance, branching pattern [9], shape irregularity and branch tortuosity [10], and blood vessel volume and density [11, 12] have all been used to describe the characteristics of vascular trees in normal organs or tumors. Second, different data parameterizations usually correspond to different topological spaces (i.e., non-homeomorphic), which can have distinct properties from each other so the theories and computing algorithms for data analysis need to be tailored for different parameterizations. Third, it has been recognized that the spaces where the tree-structured data lie in could be strongly non-Euclidean [13], the conventional analysis methods (e.g., principle component analysis) formulated in the Euclidean space are thus not applicable without reinven-tion. Particularly, the very basic measures in Euclidean space such as distance, mean and variance need to be re-defined and re-implemented for each different manifold, companied by a number of theoretical and computational difficulties.

Clustering is an important and often necessary tool in exploratory data analysis. By classifying patterns into groups in an unsupervised manner, tree-structured data clustering can provide important information for decision making in disease diagnosis, prognosis or treatment. However, tree-structured data clustering remains a challenging problem as all the three issues mentioned in the previous paragraph need to be addressed; and in comparison with the advance of data generation techniques, the development of clustering algorithm for tree-structured data has significantly lagged behind. This paper propose a novel *Topology-Attribute Matrix Parameterization* (TAMP) and a *Topology-Attribute Matrix BAsed Clustering* (TAM-BAC) framework for tree-structured data. Briefly, the topology-attribute matrix (T-A matrix) is a flexible representation of trees, which encodes the tree topology in its row indices and can accommodate an arbitrary number of attributes in its columns. The idea of topology encoding is inspired by but has evolved into a form different from the tree-line concept in [13]. More specifically, the so-called support tree is constructed such that a sub-tree can always be found in the support tree to have exactly the same topology as that of any specific tree in a sample. Each of the branches (or nodes) in the support tree is assigned a unique integer number (called the level-order index), and then the first branch (or node) will correspond to the first row of the T-A matrix, the second branch to the second row, and so forth. In this way, one T-A matrix will be generated for each single tree with its row number being equal to the total number of the support tree branches. If a branch (or node) of the support tree is missing from an individual tree, the corresponding row in the T-A matrix is simply filled with zeros; otherwise, the attributes associated with the individual tree's branch are filled into the row. Furthermore, if only the

---

- *N. Lu is with State Key Laboratory for Manufacturing Systems Engineering, Systems Engineering Institute, Xi'an Jiaotong University, Xi'an, Shaanxi,China, 710049. E-mail: lvna2009@mail.xjtu.edu.cn.*
- *H. Miao is with University of Texas Health Science Center at Houston, Department of Biostatistics, Houston, TX, USA, 77030. E-mail: Hongyu.Miao@uth.tmc.edu*

tree topology is of concern, the T-A matrix will degenerate to a vector with its element being 1 for an existing branch and 0 for a missing branch. By converting trees into matrices, the novel TAMP approach provides an intuitive way to encode both tree topology and associated attributes; also, the development of quantitative data analysis methods can now be conducted on matrix manifold, which could benefit from several pioneer studies in this field [14-16]. The underlying geometry of the T-A matrix manifold is usually non-Euclidean and complicated, so it is natural to consider the nonlinear dimensionality reduction methods such as nonnegative matrix factorization (NMF) [17], given the fact that branch attributes (e.g., length, radius, tortuosity) can only take nonnegative values. However, the classic NMF method is not suitable for the T-A matrices due to the existence of additional structure constraints on trees; for example, to be meaningful, a positive branch radius must be companied with a positive branch length, which is not accounted for by the non-negativity constraints in the traditional NMF methods. Therefore, a structure-constrained NMF (SCNMF) method is developed in this study, which utilizes a kernel function at each iteration of the matrix factorization to fulfill the structure constraints. Like the eigenvectors in PCA, a number of meta-trees obtained using the SCNMF method form the bases so each single tree in a sample can be represented by a linear combination of these metatrees. The linear coefficient vectors in front of the metatrees are the signature vectors of trees and accordingly, the tree clustering problem is solved in this signature vector space (also called meta-tree space). For this purpose, we further explore the property and metrics in the metatree space and accomplish the classification task based on the concepts like Fréchet mean.

The rest of the paper is organized as follows: the related work is described in Section 2, the details of the proposed framework are presented in Section 3, the algorithm evaluation and comparison are presented and discussed in Section 4, and this work is summarized in Section 5.

## 2 RELATED WORKS

### 2.1 Data Parameterization

Complex objects such as trees and networks are called object-oriented data, which are not directly usable in quantitative analysis without approximate parameterization. Although the parameterization problem of tree-structured data has been previously tackled [18-20], there is no a unanimously accepted method and it is still desirable to explore more flexible and intuitive ways for tree parameterization. More specifically, the classical methods such as the adjacency matrix, degree matrix, or Laplacian matrix [21] primarily focus on encoding the topological information, but leave little room to accommodate additional information like edge attributes. Feragen et al. [22] followed the convention of graph theory and used the pair $(\Gamma, x)$ to represent trees, where $\Gamma = (V, E, r)$ contains the tree topology with $V$ being the vertex set, $E$ the edge set and $r$ the root vertex, and $x$ contains edge attributes such as length or landmark points. Based on this representation, the geodesic, called the quotient Euclidean distance (QED), between two trees can be determined by searching the optimal path throughout the possible transitions from one tree to another. However, Feragen's method is intrinsically an edge-matching based method that requires both topological alignment and geometrical matching, and is thus computationally expensive due to its combinatorial essence.

Wang and Marron [13, 23, 24] proposed the tree-line idea to parameterize the tree-structured data. For a tree population, a sufficiently large tree (called support tree) was constructed and its branches were numbered, then the branches of an individual tree in the population can be lined into a vector according to their indices in the support tree. To account for both topological and geometrical information, structure tree-line and attribute tree-line were constructed separately. Accordingly, the distances had to be derived for the structure and the attribute tree-lines, respectively, and then an empirical weighted sum of the two distances was adopted as the measure of the overall distance between two trees. However, it has been claimed in [22] that this metric is discontinuous and sensitive to structural noise.

Billera et al. [19] characterized the tree-structured data in a stitched space, which is constructed by gluing multiple orthants that represent trees of different topology. The dimension of each orthant depends on the number of interior edges (that is, edges that are not connected with any terminal node), and the coordinates in each orthant reflect the length of the interior edges. However, this representation can only accommodate one geometrical feature (edge length), which is sufficient for phylogenetic tree studies but not for more general trees such as vessels and pulmonary airway structures. Owen et al. [25] developed an algorithm with a polynomial complexity to calculate the geodesic between trees in the tree space above. Nye [26] also adopted this tree representation and developed an PCA analog for phylogenetic trees. However, the computing burden associated with this representation will drastically increase as the number of the leaf nodes increases; e.g., 15 quadrants are needed for trees with 5 terminal nodes to construct the stitched space, but 512 quadrants are needed if the number of terminal nodes increases to 10.

### 2.2 Previous Analysis Methods

In recent years, increasing efforts have been dedicated to the development of more appropriate algorithms for tree-structured data analysis. Given the complex geometry of tree-structured data spaces, there has been great interest in the development of distance metrics for different parameterization. Early work such as the Nearest Neighbor Interchange (NNI) distance [18], the Robinson-Foulds distance [27], the Subtree-Prune-and-Regraft (SPR) distance [28], and the Tree Bisection and Reconnection (TBR) distance [29] can only address the topology difference and are associated with a heavy computing burden. Billera et al. [19] proposed a segmented geodesic for trees lying in different orthants; and within the same orthant, the metric could be a geodesic on a spherical triangle, called cone path. Sebastian et al. [30] described the Tree Edit distance (TED), which is the minimal cost of transforming one tree to another by adding or removing edges. Both the segmented cone path and the TED are computationally expensive; also, the geodesic between trees is not unique based on the TED definition, which is problematic when calculating the mean or variance of shapes. Wang and Marron [13] defined an empirical metric for binary trees, which has an integer part for topological difference and a fractional part for geometrical difference. Feragen et al. [22, 31] proposed the Quotient Euclidean distance (QED), which is the concatenations of straight lines in the quotient space of equivalent trees. Like the other metrics mentioned before, the computing complexity of QED is also of concern. In addition, one common issue of all the metrics above is that they have a limited capacity to accommodate more than one attribute [19, 26, 32].

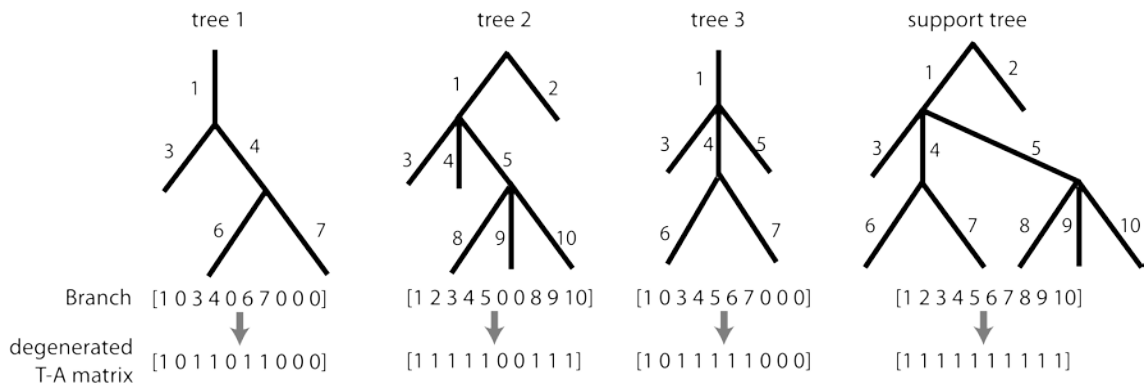

Fig. 1. Illustration of the support tree, the branch indexing scheme, and the degenerated T-A matrix.

Based on distance metrics, several recent studies tackled the problem of calculating the basic statistics in the tree data spaces, e.g., mean (centroid, or midpoint) and variation [33]. For instance, Feragen et al. [33] suggested three iterative methods for the calculation of the tree mean (that is, centroids, Birkhoff shortening and weighted midpoints). Billera et al. [19] proposed an alternative way to obtain the centroid of trees by iteratively reaching the midpoint via a converging sequence of the geodesics between trees. Mean and variation calculation make it feasible to develop more advanced analysis approaches for tree-structured data. Wang, Aydin and their colleagues [13, 24, 34] proposed an analog of PCA, which determines the principle tree-line components based on the projection of trees onto tree-lines. However, this line of ideas is essentially empirical due to the lack of a formal guideline for the choice of weights on topological and geometrical information. Nye [26] proposed an alternative analog of PCA in the CAT(0) metric space of non-positive curvature. However, this approach was primarily designed to investigate phylogenetic trees such that it is not straightforward to extend the method to accommodate more attributes other than edge length. Also, the search for the optimal simple line, which is the analog of the principle component in the regular PCA, could be computationally demanding for complex tree structures. Nevertheless, all the researches mentioned above have made inspiring attempts to develop useful quantitative methods for tree analysis.

## 3 METHODS

The details of the TAMBAC algorithm based on T-A matrix parameterization are described in this section. For now on, we only consider trees that have geometric attributes associated with their edges for simplicity; however, the idea is applicable to general tree-structured data.

### 3.1 Topology-Attribute Matrix

In this section, we illustrate how to represent a tree using the topology-attribute matrix (T-A matrix). For a given population of trees, let $d_{max}$ be the largest depth of these trees and $m$ be the maximum number of branches (also called tree order) that a tree vertex can have, a support tree (also called maximal tree [35]) is then an $m$-ary tree [36] with the depth $d_{max}$. Assume that the root vertex of the support tree is on the top, then all the branches of the support tree are consecutively numbered in an ascending manner from top to bottom and from left to right. For any individual tree in the given population, each of its branches will correspond to only one branch of the support tree by definition, and thus acquires the index of the support tree branch. Now let $p$ denote the number of branches in the support tree (called the dimension of the support tree), a $p$-dimensional vector can be constructed for each individual tree in a way such that the $i$-th vector element is assigned the value 1 if the $i$-th branch of the support tree is also contained in the individual tree; otherwise, zero is assigned to the $i$-th element. Such a vector is actually the degenerated T-A matrix if only tree topology is of concern. See Fig. 1 for a schematic illustration of the construction of the support tree, the branch indexing strategy and the degenerated T-A matrices based on a population of three trees.

The degenerated T-A matrix can be easily extended to accommodate additional tree attributes other than topology. As shown in Fig. 2, for the degenerated T-A matrix of tree 1 in Fig. 1, if all the ones are replaced with the geometric attribute (e.g., edge length) of the corresponding branches, the degenerated T-A matrix now encodes both topology and one attribute. To accommodate multiple attributes such as branch radius and tortuosity, one can simply add more columns to the T-A matrix, and fill zeros in the rows corresponding to non-exist branches (Fig. 2). One can immediately tell that the way in which the T-A matrix encodes information is applicable to general trees, and it provides an intuitive representation in the sense that the tree topology is encoded in row indices and the attributes are explicitly shown in different columns.

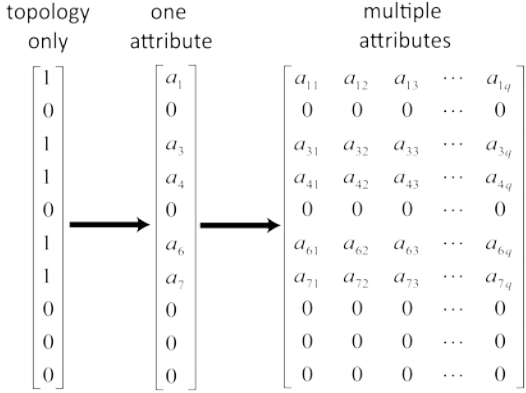

Fig. 2. Examples of the degenerated T-A matrix, T-A matrix with one attribute, and T-A matrix with multiple attributes.

## 3.2 Structure Constrained NMF

For a population of trees $\{t_1, t_2, \ldots, t_n\}$, let $s$ denote the support tree with a total of $p$ branches. If $q$ attributes are of concern, then a T-A matrix $\mathbf{T}_{p \times q}$ can be generated for each tree. With this representation, tree clustering can be conducted in the matrix space $\mathcal{M}$ expanded by $\{\mathbf{T}_1, \mathbf{T}_2, \ldots, \mathbf{T}_n\}$. For this purpose, a feasible strategy is to identify the bases of $\mathcal{M}$ first, then each $\mathbf{T}_i$ ($i$=1,2,…$n$) can be decomposed into a linear combination of these bases such that the similarity (or dissimilarity) between a pair of trees can be defined based on the coefficients of the linear combination. Considering that the geometric attributes associated with tree branches like length, radius and tortuosity are all non-negative, it is natural to consider the non-negative matrix factorization (NMF) [17] when identifying the bases of $\mathcal{M}$. There are two more reasons that make the use of NMF attractive: first, NMF does not make the orthogonality assumption as in PCA and is thus applicable to a broad range of matrix manifolds; second, the bases obtained using NMF are physically meaningful parts [37], called *meta-tree* in the context of this study. However, the traditional NMF method cannot be directly applied to tree-structured data due to the existence of additional constraints other than non-negativity. More specifically, it is impossible for a real tree branch to have a positive radius but a length of zero; that is, all the entries of the same row in a T-A matrix must be simultaneously non-zero or zero, which imposes a structure constraint on T-A matrices. *As a matter of fact, structure con-*

*straints also exist in other object-oriented data like face images and have to be considered to extract meaningful intrinsic parts. This observation necessitates the development of the structure-constrained NMF (SCNMF).*

The manifold $\mathcal{M}$ has a natural chart $\varphi : \mathbf{T}_{p \times q} \to t_{pq \times 1}$, where $t_{pq \times 1}$ denotes a $pq$-dimensional vector obtained by stacking the columns of $\mathbf{T}$ one above another. This chart enables us to define the so-called forest matrix $\mathbf{F} = [t_1, \cdots, t_n]$, and the SCNMF seeks to decompose this matrix as follows

$$\mathbf{F}_{pq \times n} \approx \tau\left(\mathbf{W}_{pq \times k} \cdot \mathbf{H}_{k \times n}\right), \tag{1}$$

where $\tau(\cdot)$ is a map that imposes the structure constraint on the product $\mathbf{W} \cdot \mathbf{H}$. Note that the columns of $\mathbf{W}$ are actually the meta-trees mentioned above, and the columns of $\mathbf{H}$ are the coefficients of the linear combination of these meta-trees. Based on this interpretation, the structure constraint can be simply imposed on $\mathbf{W}$ and the SCNMF becomes

$$\mathbf{F}_{pq \times n} \approx \tau\left(\mathbf{W}\right)_{pq \times k} \cdot \mathbf{H}_{k \times n}. \tag{2}$$

To accomplish the factorization above, the following Lagrange function is considered

$$\mathcal{L} = \left\|\mathbf{F} - \tau\left(\mathbf{W}\right) \cdot \mathbf{H}\right\|^2 + \mathrm{Tr}\left(\mathbf{A} \cdot \tau\left(\mathbf{W}\right)^T\right) + \mathrm{Tr}\left(\mathbf{B}\mathbf{H}^T\right), \tag{3}$$

where $\|\cdot\|$ denotes the Frobenius norm (or called the Hilbert–Schmidt norm), and $\mathbf{A}_{pq \times k}$ and $\mathbf{B}_{k \times n}$ are the Lagrange multipliers for the non-negativity constraints on $\tau(\mathbf{W})$ and $\mathbf{H}$, respectively. One can show that the minima of $\mathcal{L}$ correspond to the solutions of the following equations

$$\frac{\partial \mathcal{L}}{\partial \tau(\mathbf{W})} = -2\mathbf{F}\mathbf{H}^T + 2\tau(\mathbf{W}) \cdot \mathbf{H}\mathbf{H}^T + \mathbf{A} = \mathbf{0}, \tag{4}$$

$$\frac{\partial \mathcal{L}}{\partial \mathbf{H}^T} = -2\mathbf{F}^T \cdot \tau(\mathbf{W}) + 2\mathbf{H}^T \cdot \tau(\mathbf{W})^T \cdot \tau(\mathbf{W}) + \mathbf{B}^T = \mathbf{0}. \tag{5}$$

Considering the Karush-Kuhn-Tucker (KKT) conditions $\mathbf{A}_{(i,i)}\tau(\mathbf{W})_{(j,i)} = 0$ and $\mathbf{B}_{(j,i)}\mathbf{H}_{(i,i)} = 0$, one can obtain

$$-\left(\mathbf{F}\mathbf{H}^T\right)_{(i,i)} \cdot \tau(\mathbf{W})_{(j,i)} + \left(\tau(\mathbf{W}) \cdot \mathbf{H}\mathbf{H}^T\right)_{(i,i)} \cdot \tau(\mathbf{W})_{(j,i)} = 0,$$

$$-\left(\mathbf{F}^T \cdot \tau(\mathbf{W})\right)_{(i,i)} \cdot \mathbf{H}_{(i,i)} + \left(\mathbf{H}^T \cdot \tau(\mathbf{W})^T \cdot \tau(\mathbf{W})\right)_{(i,i)} \cdot \mathbf{H}_{(i,i)} = 0,$$

where the subscript $\square_{(i,i)}$ denotes the matrix element at the $i$-th row and the $j$-th column. Based on the equations above, the updating rules can be obtained as follows

$$\tau(\mathbf{W})_{(i,i)} \leftarrow \tau(\mathbf{W})_{(i,i)} \frac{\left(\mathbf{F}\mathbf{H}^T\right)_{(j,i)}}{\left(\tau(\mathbf{W}) \cdot \mathbf{H}\mathbf{H}^T\right)_{(j,i)}}, \tag{6}$$

$$\mathbf{H}_{(i,j)} \leftarrow \mathbf{H}_{(i,j)} \frac{\left(\mathbf{F}^T \cdot \tau(\mathbf{W})\right)_{(i,j)}}{\left(\mathbf{H}^T \cdot \tau(\mathbf{W})^T \cdot \tau(\mathbf{W})\right)_{(i,j)}} . \tag{7}$$

The key difference between the SCNMF updating rules and the classic NMF updating rules is the introduction of the function $\tau(\cdot)$ into Eqns. (6) and (7) to impose the structure constraints on $\mathbf{W}$. Here we propose an implementation of such a function to assure that the entries of a T-A matrix row are all positive or all zeros (called *positive uniformity constraint*) as follows

$$\tau(\mathbf{W})_{\cdot j} = \mathbf{W}_{\cdot j} + \rho\left(\lambda \cdot g\left(\varphi^{-1}\left(\mathbf{W}_{\cdot j}\right) . \wedge \mathbf{1}_{p \times q}\right)\right), \tag{8}$$

where the subscript $\square_{\cdot j}$ denotes the $j$-th column of a matrix, $\mathbf{1}_{p \times q}$ denotes a matrix with all its elements being 1, $.\wedge$ is an element-wise "AND" operation, and $\varphi^{-1}(\cdot)$ is the inverse chart that transforms the $j$-th column of $\mathbf{W}$ back into a $p \times q$ T-A matrix. Furthermore, $g(\cdot)$ stands for the operation of matrix row summation followed by thresholding, which checks whether the entries in a row have the positive uniformity; $\lambda$ is a small positive value specified by users to control the correction of the row entries that violate the positive uniformity; and $\rho(\cdot)$ is a reshape function that repeats a $p$-dimensional vector for $q$ times to form a matrix column of length $pq$.

It is also necessary to show that the SCNMF approach is locally convergent since the NMF-type methods do not have a unique solution in general. Equivalently, we prove the following theorem.

**Theorem 1.** The objective function $O = \left\| \mathbf{F} - \tau(\mathbf{W}) \cdot \mathbf{H} \right\|^2$ is non-increasing under the update rules in Eqns. (6) and (7).
**Proof.**
Define the function

$$\Psi(\mathbf{H}_{\cdot l}) = \frac{1}{2}\left(\mathbf{F}_{\cdot l} - \tau(\mathbf{W}) \cdot \mathbf{H}_{\cdot l}\right)^T \left(\mathbf{F}_{\cdot l} - \tau(\mathbf{W}) \cdot \mathbf{H}_{\cdot l}\right),$$
$$l = 1, 2, ..., n , \tag{9}$$

and one can tell $O = 2\sum_{l=1}^{n} \Psi(\mathbf{H}_{\cdot l})$ by definition of the Frobenius norm. Therefore, if we can show that each $\Psi(\mathbf{H}_{\cdot l})$ is non-increasing under the update rules, then so is $O$.

For this purpose, the following function can be defined

$$G\left(\mathbf{H}_{\cdot l}, \mathbf{H}_{\cdot l}^{\hat{k}}\right) = \Psi\left(\mathbf{H}_{\cdot l}^{\hat{k}}\right) + \left(\mathbf{H}_{\cdot l} - \mathbf{H}_{\cdot l}^{\hat{k}}\right)^T \cdot \nabla\Psi\left(\mathbf{H}_{\cdot l}^{\hat{k}}\right)$$
$$+ \frac{1}{2}\left(\mathbf{H}_{\cdot l} - \mathbf{H}_{\cdot l}^{\hat{k}}\right)^T \cdot \mathbf{D}\left(\mathbf{H}_{\cdot l}^{\hat{k}}\right) \cdot \left(\mathbf{H}_{\cdot l} - \mathbf{H}_{\cdot l}^{\hat{k}}\right), \tag{10}$$

where $\mathbf{H}_{\cdot l}^{\hat{k}}$ denotes the $l$-th column of the $\hat{k}$-th matrix in a sequence of matrices, and $\mathbf{D}$ is a diagonal matrix given by

$$\mathbf{D}_{(i,j)}(\mathbf{H}_{\cdot l}^{\hat{k}}) = \delta_{ij}\left(\tau(\mathbf{W})^T \cdot \tau(\mathbf{W}) \cdot \mathbf{H}_{\cdot l}^{\hat{k}}\right)_{(i)} \Big/ \mathbf{H}_{(i,l)}^{\hat{k}} .$$

According to Lemma 2 in [38], $G\left(\mathbf{H}_{\cdot l}, \mathbf{H}_{\cdot l}^{\hat{k}}\right)$ is the so-called auxiliary function for $\Psi(\mathbf{H}_{\cdot l})$ such that $G\left(\mathbf{H}_{\cdot l}, \mathbf{H}_{\cdot l}^{\hat{k}}\right) \ge \Psi(\mathbf{H}_{\cdot l})$ and $G\left(\mathbf{H}_{\cdot l}, \mathbf{H}_{\cdot l}\right) = \Psi(\mathbf{H}_{\cdot l})$. By Lemma 1 in [38], we know that $\Psi(\mathbf{H}_{\cdot l})$ is non-increasing as $\hat{k}$ increases if $G\left(\mathbf{H}_{\cdot l}, \mathbf{H}_{\cdot l}^{\hat{k}}\right)$ is an auxiliary function and

$$\mathbf{H}_{\cdot l}^{\hat{k}+1} = \arg\min_{\mathbf{H}_{\cdot l}} G(\mathbf{H}_{\cdot l}, \mathbf{H}_{\cdot l}^{\hat{k}}) . \tag{11}$$

Now we only need to show that Eq. (11) is equivalent to the update rule in Eq. (7). Note that $\mathbf{H}_{\cdot l}^{\hat{k}+1}$ is the solution of the following equation

$$\frac{\partial}{\partial \mathbf{H}_{\cdot l}} G(\mathbf{H}_{\cdot l}, \mathbf{H}_{\cdot l}^{\hat{k}})\bigg|_{\mathbf{H}_{\cdot l} = \mathbf{H}_{\cdot l}^{\hat{k}+1}} = 0$$
$$\Rightarrow \nabla\Psi\left(\mathbf{H}_{\cdot l}^{\hat{k}}\right) + \mathbf{D}\left(\mathbf{H}_{\cdot l}^{\hat{k}}\right) \cdot \left(\mathbf{H}_{\cdot l}^{\hat{k}+1} - \mathbf{H}_{\cdot l}^{\hat{k}}\right) = 0$$
$$\Rightarrow \mathbf{H}_{\cdot l}^{\hat{k}+1} = \mathbf{H}_{\cdot l}^{\hat{k}} - \mathbf{D}(\mathbf{H}_{\cdot l}^{\hat{k}})^{-1} \cdot \nabla\Psi(\mathbf{H}_{\cdot l}^{\hat{k}})$$
$$\Rightarrow \mathbf{H}_{(i,l)}^{\hat{k}+1} = \mathbf{H}_{(i,l)}^{\hat{k}} - \frac{\mathbf{H}_{(i,l)}^{\hat{k}}}{\left(\tau(\mathbf{W})^T \cdot \tau(\mathbf{W}) \cdot \mathbf{H}_{\cdot l}^{\hat{k}}\right)_{(i)}}$$
$$\cdot \left[\left(\tau(\mathbf{W})^T \cdot \tau(\mathbf{W}) \cdot \mathbf{H}_{\cdot l}^{\hat{k}}\right)_{(i)} - \left(\tau(\mathbf{W})^T \cdot \mathbf{F}_{\cdot l}\right)_{(i)}\right]$$
$$\Rightarrow \mathbf{H}_{(i,l)}^{\hat{k}+1} = \mathbf{H}_{(i,l)}^{\hat{k}} \frac{\left(\mathbf{F}^T \cdot \tau(\mathbf{W})\right)_{(i,l)}}{\left(\left(\mathbf{H}^{\hat{k}}\right)^T \cdot \tau(\mathbf{W})^T \cdot \tau(\mathbf{W})\right)_{(i,l)}},$$

which completes the proof. ∎

### 3.3 Property and Metric of Meta-tree Space

After matrix factorization, the forest matrix $\mathbf{F}$ is decomposed into the product of $\tau(\mathbf{W})$ and $\mathbf{H}$. Let $\boldsymbol{m}_i$ ($i = 1, 2, ..., k$) denote the $i$-th column of $\tau(\mathbf{W})$; then a column of $\mathbf{H}$, which is called the *signature vector* and denoted by $\boldsymbol{h}_j$ ($j = 1, 2, ..., n$), is the coordinate of a tree in the meta-tree space $\mathcal{H}$ spanned by $\{\boldsymbol{m}_1, \boldsymbol{m}_2, ..., \boldsymbol{m}_k\}$. Of course, if we rotate trees or index tree branches differently, we will usually get different $\tau(\mathbf{W})$ and $\mathbf{H}$, and thus a different but equivalent meta-tree space $\mathcal{H}$. For simplicity as in [22], here we assume all trees have been appropriately aligned and all tree branches are numbered using the same rule (e.g., the one described in Section 3.1).

One can tell that the meta-tree space $\mathcal{H}$ is a specific type of topological space called *cone* [39], which has the following properties:

**P1.** $\mathcal{H}$ is closed, nonempty and $\mathcal{H} \ne \{0\}$;

**P2.** If $a, b \in \mathbb{R}$, $a, b \ge 0$ and $\boldsymbol{h}_1, \boldsymbol{h}_2 \in \mathcal{H}$, then $a\boldsymbol{h}_1 + b\boldsymbol{h}_2 \in \mathcal{H}$;

**P3.** If there exist $\boldsymbol{h} \in \mathcal{H}$ and $-\boldsymbol{h} \in \mathcal{H}$, then $\boldsymbol{h}$ is a zero vector.

Also, since every meta-tree $\boldsymbol{m}_i$ is nonnegative, the inner product $\boldsymbol{m}_i \cdot \boldsymbol{m}_j$ ($i \neq j$) is not necessarily equal to zero so the basis vectors $\{\boldsymbol{m}_1, \boldsymbol{m}_2, ..., \boldsymbol{m}_k\}$ are usually not orthogonal to each other. Actually, the meta-tree space $\mathcal{H}$ has a skew coordinate system (see Fig. 3 for illustration, where each axis is a $pq$-dimensional vector); therefore, the regular Euclidean distance (that is, the $L^2$ norm of $\boldsymbol{h}_i - \boldsymbol{h}_j$) is not the geodesic between two points in the meta-tree space and may not perform well if used for the clustering task. We thus need to further explore the metrics in the meta-tree space.

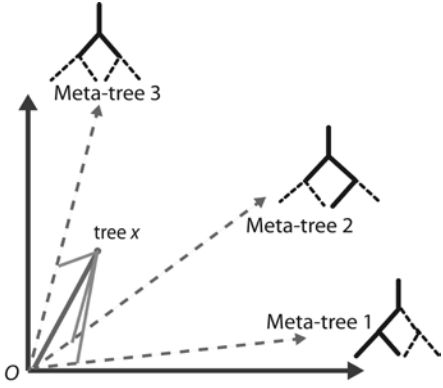

Fig. 3. Illustration of the coordinate system of a meta-tree space. The dashed axes represent the basis vectors spanning the meta-tree space and the solid-line axes are for an orthogonal coordinate system.

The primary difficulty in defining distances between trees is that there may exist an infinite number of paths between trees if we consider all possible continuous transformations; and even if we only consider discrete transformations as in the conventional tree edit operation, the number of paths could still be very large given the combinatorial nature of such an operation. For illustration purpose, Fig. 4 shows two possible transformations from tree $\boldsymbol{t}_1$ to tree $\boldsymbol{t}_2$, and one can calculate the quotient Euclidean distances [20] for the two paths in the original tree space as $\sqrt{2e_4{}^2 + e_5{}^2 + (e_5 - e_2)^2} + \sqrt{2e_3{}^2 + e_5{}^2 + (e_5 - e_2)^2}$ and $\sqrt{2(e_3{}^2 + e_4{}^2)}$, respectively. However, in practice, it is computationally expensive or even unaffordable to calculate the distance of every possible path and then find out the shortest one(s). Unfortunately, we have the same problem in the meta-tree space, considering the fact that any path between two points $\boldsymbol{h}_1$ and $\boldsymbol{h}_2$ in the meta-tree space must correspond to a feasible transformation from tree $\boldsymbol{t}_1$ to $\boldsymbol{t}_2$. Currently, techniques like semi-labeling and approximation have been employed to enumerate all paths and compute the associated QEDs for small trees [22]; however, such approaches may not be efficient for general trees and are not directly applicable to the meta-tree space. Therefore, instead of geodesics like QED in the original tree space, here we seek for a computationally-efficient and simple-to-implement approximation to the exact geodesic in the meta-tree space, based on the concept of strict consensus tree [40-43]. By definition, a strict

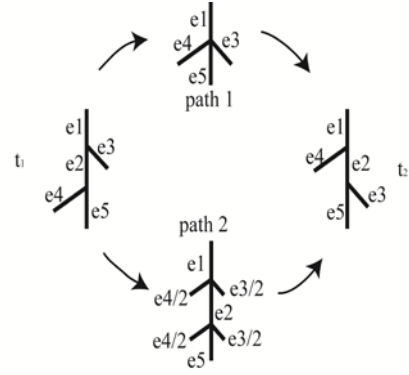

Path 1: [e1 0 e2 e3 e4 e5 0]→[e1 e4 e3 0 0 0]→[e1 e4 e2 0 0 e5 e3]
Path 2: [e1 0 e2 e3 e4 e5 0]→[e1 e4/2 e2 e3/2 e4/2 e5 e3/2]→[e1 e4 e2 0 0 e5 e3]

Fig. 4. Illustration of transforms between two trees.

consensus tree of two trees only contains edges that are in both trees. Amenta et al. [44] have shown that in the original tree space, the length of the path through the strict consensus tree is the upper bound of the tree geodesic distance, and this upper bound is at most $\sqrt{2}$ times greater than the lower bound of the geodesic. Borrowing the idea in [44] to the meta-tree space, we show that the $L^1$ norm can be used to calculate the length of the path through the strict consensus tree, and the $L^1$ norm is at most 2 times greater than the lower bound of the geodesic. Therefore, we can use the $L^1$ norm as the distance between two signature vectors in the meta-tree space, which is simple-to-implement and turns out to outperform both the $L^2$ norm in the meta-tree space and other two state-of-the-art metrics in the original tree space (see Section 4.4

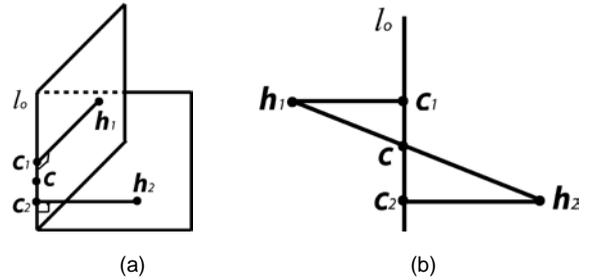

Fig. 5. The transition path through a strict consensus tree in the meta-tree space. (a) Tree transition path, (b) Unfolded transition path.

for details).

We need to define the path through the strict consensus tree in the meta-tree space first. For convenience, the signature vectors in the meta-tree space are still called "trees" from now on. According to the definition of the meta-tree space $\mathcal{H}$, we know that all trees residing on the same line that passes through the origin (denoted by $l_o$) will have the same topology but different geometric attributes. Thus, to determine the path through the strict consensus tree between trees $\boldsymbol{h}_1$ and $\boldsymbol{h}_2$, we can shrink them to two other trees ($\boldsymbol{c}_1$ and $\boldsymbol{c}_2$) that share the same topology by projecting $\boldsymbol{h}_1$ and $\boldsymbol{h}_2$ to one line passing through the origin in the meta-tree space. Fig. 5(a) visualizes such a path through a strict consensus tree **c** that resides on line $l_o$, and the unfolded version of this path into a plane is shown in Fig. 5(b). The length of this path is the upper bound of the geodesic distance in the meta-tree space [44], which can be calculated as follows

$$d_{L2}(\mathbf{h}_1, \mathbf{h}_2) = \sqrt{d(\mathbf{h}_1, \mathbf{c}_1)^2 + d(\mathbf{c}_1, \mathbf{c})^2} + \sqrt{d(\mathbf{c}, \mathbf{c}_2)^2 + d(\mathbf{c}_2, \mathbf{h}_2)^2} \ , \quad (12)$$

where $d(\cdot, \cdot)$ denotes the distance between two trees in the same plane.

In the calculation above, one needs to find the strict consensus tree first, which may not be a trivial computing task. Therefore, a computationally-efficient metric alternative to the distance defined in Eq. (12) is desirable. For this purpose, we consider an upper bound of $d_{L2}$ as follows

$$d_{UB} = d(\mathbf{h}_1, \mathbf{c}_1) + d(\mathbf{c}_1, \mathbf{c}) + d(\mathbf{c}, \mathbf{c}_2) + d(\mathbf{c}_2, \mathbf{h}_2) \geq d_{L2}(\mathbf{h}_1, \mathbf{h}_2) \ . \quad (13)$$

Now if $d(\cdot, \cdot)$ is calculated using the $L^1$ norm, we can define the following metric based on Eq. (13) as

$$\begin{aligned} d_{L1} &= d_{UB} \\ &= d(\mathbf{h}_1, \mathbf{c}_1) + d(\mathbf{c}_1, \mathbf{c}) + d(\mathbf{c}, \mathbf{c}_2) + d(\mathbf{c}_2, \mathbf{h}_2) \\ &= |\mathbf{h}_1 - \mathbf{c}_1| + |\mathbf{c}_1 - \mathbf{c}| + |\mathbf{c} - \mathbf{c}_2| + |\mathbf{c}_2 - \mathbf{h}_2| \quad (\text{if } d(\cdot) = |\cdot|) \\ &= |\mathbf{h}_1 - \mathbf{h}_2|. \end{aligned} \quad (14)$$

One can immediately tell that $d_{L1}$ is just the $L^1$ norm along the path through the strict consensus tree in the meta-tree space. Interestingly, we can show that $d_{L1}$ differs from the lower bound of the geodesic at most by a factor of 2.

**Theorem 2.** In the meta-tree space, $d_{L1}$ differs from the lower bound of the geodesic at most by a factor of 2.

**Proof.** The ratio between $d_{L1}$ and $d_{L2}$ is

$$R = \frac{d_{L1}}{d_{L2}} = \frac{d(\mathbf{h}_1, \mathbf{c}_1) + d(\mathbf{c}_1, \mathbf{c}) + d(\mathbf{c}, \mathbf{c}_2) + d(\mathbf{c}_2, \mathbf{h}_2)}{\sqrt{d(\mathbf{h}_1, \mathbf{c}_1)^2 + d(\mathbf{c}_1, \mathbf{c})^2} + \sqrt{d(\mathbf{c}, \mathbf{c}_2)^2 + d(\mathbf{c}_2, \mathbf{h}_2)^2}}$$

For convenience, denote $a_1 = d(\mathbf{h}_1, \mathbf{c}_1)$ , $b_1 = d(\mathbf{c}_1, \mathbf{c})$ , $a_2 = d(\mathbf{c}, \mathbf{c}_2)$ and $b_2 = d(\mathbf{c}_2, \mathbf{h}_2)$ , then we have

$$R = \frac{a_1 + b_1 + a_2 + b_2}{\sqrt{a_1^2 + b_1^2} + \sqrt{a_2^2 + b_2^2}} \ .$$

It is easy to verify that $R$ reaches its maximum value $\sqrt{2}$ when $a_1 = b_1$ and $a_2 = b_2$. That is,

$$d_{L1} \leq \sqrt{2} d_{L2} \ .$$

It has been shown by Theorem 2 in [44] that $d_{L2}$ is at most $\sqrt{2}$ times greater than the lower bound of the geo-

desic. Let $g_{LB}$ denote the lower bound of the geodesic, we then have

$$d_{L1} \leq \sqrt{2} d_{L2} \leq \sqrt{2} \cdot \sqrt{2} \cdot g_{LB} = 2 \cdot g_{LB} \ ,$$

which completes the proof. ∎

Theorem 2 suggests $d_{L1}$ can be used as an approximation to the geodesic in the meta-tree space and one can also tell that this metric is simple to implement. We conduct extensive experiment studies in Section 4 and validate the performance of this geodesic approximation.

### 3.4 Tree Clustering in the Meta-tree Space

Once the geodesic distance is defined in the meta-tree space, different clustering methods can be incorporated into the TAMBAC framework. For evaluation purpose, we consider two widely-used clustering methods, K-means and NCut [45], which utilize the distance metric differently. More specifically, NCut uses the distance metric in the meta-tree space to calculate the weighted graph, while K-means calculates the mean tree based on the distance metric. Since the meta-tree space is not a standard Euclidean space, particular attention needs to be paid when calculating the coordinates of mean tree for the K-means method. By definition, the coordinates of Fréchet mean tree is given as follows:

$$\mu = \arg\min_{\mathbf{h}} \sum_{i=1}^{n} d(\mathbf{h}, \mathbf{h}_i)^2 \ ,$$

where $d(\cdot)$ is a distance metric and $\mathbf{h}_i \ (i = 1, 2, \ldots, n)$ are the coordinates of trees in the meta-tree space. As suggested in [33], the weighted midpoint method is used to approximate the Fréchet mean in this study, which takes the following form

$$\mu(\{\mathbf{h}_1, \mathbf{h}_2, \ldots, \mathbf{h}_n\}) = \frac{\mathbf{h}_i + (n-1)\mu(\{\mathbf{h}_1, \mathbf{h}_2, \ldots, \mathbf{h}_n\} \setminus \{\mathbf{h}_i\})}{n} \ .$$

---

**Algorithm 1:** TAMBAC pipeline for tree clustering.

**Input:** Vascular tree images
**Output:** Tree membership
1: Preprocess vascular tree images
2: Skeletonize vascular trees
3: Extract geometric attributes of tree branches
4: Generate T-A matrices
5: Decompose the forest matrix using SCNMF to obtain the meta-trees and tree coordinates
6: Cluster trees using methods like NCut or the Fréchet-mean-based K-means

---

The pipeline of the TAMBAC framework for tree clustering is sketched in **Algorithm 1**. The input data are vascular tree images. Several pre-processing steps (e.g., denoising, binarizing, and gap filling [46, 47]) are applied to these images before the medial axis thinning algorithm [48] is employed to extract the vessel tree skeleton. The skeletonization results enable us to extract tree branches by detecting the junctions and end points in the skeleton using the pixel connection property. Three geometric attributes (radius, length and tortuosity) are then extracted for each tree branch using distance transform [49] or by counting the skeleton pixels. The T-A matrices of the trees are then constructed as described in Section 3.1. The SCNMF is then applied to the forest matrix **F** and the coordinates of each tree in the meta-tree space can be obtained. Finally, the clustering method (NCut or Fréchet-mean-based K-means) is employed to accomplish the learning task.

## 4 RESULTS

Extensive simulation studies and real data experiments are conducted in this section to evaluate the performance of the proposed TAMBAC framework. In addition, the performance of the proposed geodesic distance is compared with those of two state-of-the-art metrics, QED [22] and Torsello's metric [50]. Also, three geometric attributes of tree branches (length, radius and tortuosity) are considered in our studies unless explicitly stated otherwise.

### 4.1 Performance on Simulated T-A Matrices

T-A matrices in reality can be generated from, e.g., angiographic images after a number of pre-processing steps such as segmentation [46] and skeletonization [51]. However, we start from the simplest case by directly generating the T-A matrices so the errors introduced by image pre-processing can be ruled out when evaluating the algorithm performance. More complicated simulation studies and real data experiments are presented in the next two sub-sections.

To account for the possible diversity in tree topology and geometric attributes, we considered different tree orders (2-ary tree or 3-ary tree), depths (level of hierarchy), branching patterns (number of branches) and geometric features (length, radius, and tortuosity) when generating the T-A matrices. More specifically, the tree depth can be either a fixed number or a random number drawn from a binomial distribution $B[n,p]$, where $n$ is the total number of trials and $p$ is the probability of success for each trial. Three branching patterns for a node in a 2-ary tree (left only, right only, left & right) and eight branching patterns for a node in a 3-ary tree (left only, middle only, right only, left & middle, middle & right, left & right, left & middle & right) are considered. Finally, the three geometric attributes can either take a constant value or be a random value drawn from a uniform distribution $U[a,b]$. The clustering results for 2-ary trees are summarizes in Table 1. Six groups of data are generated, each consisting of three sub-groups, and each sub-group contains 100 datasets. Within each dataset, two sets of trees are generated with 10 trees in each set. The tree depth is chosen as either 3 or 5; the branching pattern is designed to be "same", "different", or "random"; and the geometric attribute is drawn from U[2,5] versus U[4,7], U[710] or U[10, 15]. Moreover, the label "*same*" for branching patterns means that all the 20 trees in the same dataset have the same branching pattern; "*different*" means that the 10 trees in one set have the same branching pattern but different from that of the 10 trees in the other set; "*random*" means that the branching pattern of each of the 20 trees is randomly selected from the three possible patterns for 2-ary trees. The clustering accuracy of each sub-group is averaged over the 100 datasets within it. The results in Table 1 clearly suggest that the TAMBAC framework performs reasonably well: for the majority of the cases, the accuracy of TAMBAC is greater than 90%. For the first case of group 1, the accuracies of TAMBAC are 71% and 64% for NCut and K-means, respectively, which is reasonable since the trees in this case only differ in geometric attributes. Also, we notice that TAMBAC with NCut can usually achieve a better accuracy than TAMBAC with K-means.

TABLE 1
PERFORMANCE OF TAMBAC ON 2-ARY TREES

| Data group | Tree order | Tree depth | Branching pattern | Geometric attribute | TAMBAC with NCut accuracy | TAMBAC with K-means accuracy |
|---|---|---|---|---|---|---|
| 1 | 2 vs 2 | 3 vs 3 | Same | U[2,5] vs U[4,7] | 0.71±0.11 | 0.64±0.10 |
| | 2 vs 2 | 3 vs 3 | Same | U[2,5] vs U[7,10] | 0.95±0.10 | 0.98±0.06 |
| | 2 vs 2 | 3 vs 3 | Same | U[2,5] vs U[10,15] | 0.99±0.03 | 1.00±0.01 |
| 2 | 2 vs 2 | 3 vs 3 | Different | U[2,5] vs U[4,7] | 1.00±0.01 | 0.98±0.05 |
| | 2 vs 2 | 3 vs 3 | Different | U[2,5] vs U[7,10] | 1.00±0.02 | 0.98±0.05 |
| | 2 vs 2 | 3 vs 3 | Different | U[2,5] vs U[10,15] | 1.00±0.01 | 0.99±0.03 |
| 3 | 2 vs 2 | 3 vs 3 | Random | U[2,5] vs U[4,7] | 0.97±0.07 | 0.95±0.11 |
| | 2 vs 2 | 3 vs 3 | Random | U[2,5] vs U[7,10] | 1.00±0.01 | 0.98±0.07 |
| | 2 vs 2 | 3 vs 3 | Random | U[2,5] vs U[10,15] | 1.00±0.00 | 0.99±0.04 |
| 4 | 2 vs 2 | 3 vs 5 | Same | U[2,5] vs U[4,7] | 0.99±0.04 | 0.96±0.06 |
| | 2 vs 2 | 3 vs 5 | Same | U[2,5] vs U[7,10] | 1.00±0.03 | 0.99±0.02 |
| | 2 vs 2 | 3 vs 5 | Same | U[2,5] vs U[10,15] | 1.00±0.00 | 0.99±0.02 |
| 5 | 2 vs 2 | 3 vs 5 | Different | U[2,5] vs U[4,7] | 1.00±0.00 | 1.00±0.00 |
| | 2 vs 2 | 3 vs 5 | Different | U[2,5] vs U[7,10] | 1.00±0.00 | 1.00±0.00 |
| | 2 vs 2 | 3 vs 5 | Different | U[2,5] vs U[10,15] | 1.00±0.00 | 1.00±0.00 |
| 6 | 2 vs 2 | 3 vs 5 | Random | U[2,5] vs U[4,7] | 0.78±0.11 | 0.74±0.11 |
| | 2 vs 2 | 3 vs 5 | Random | U[2,5] vs U[7,10] | 0.93±0.11 | 0.85±0.13 |
| | 2 vs 2 | 3 vs 5 | Random | U[2,5] vs U[10,15] | 0.98±0.07 | 0.89±0.13 |

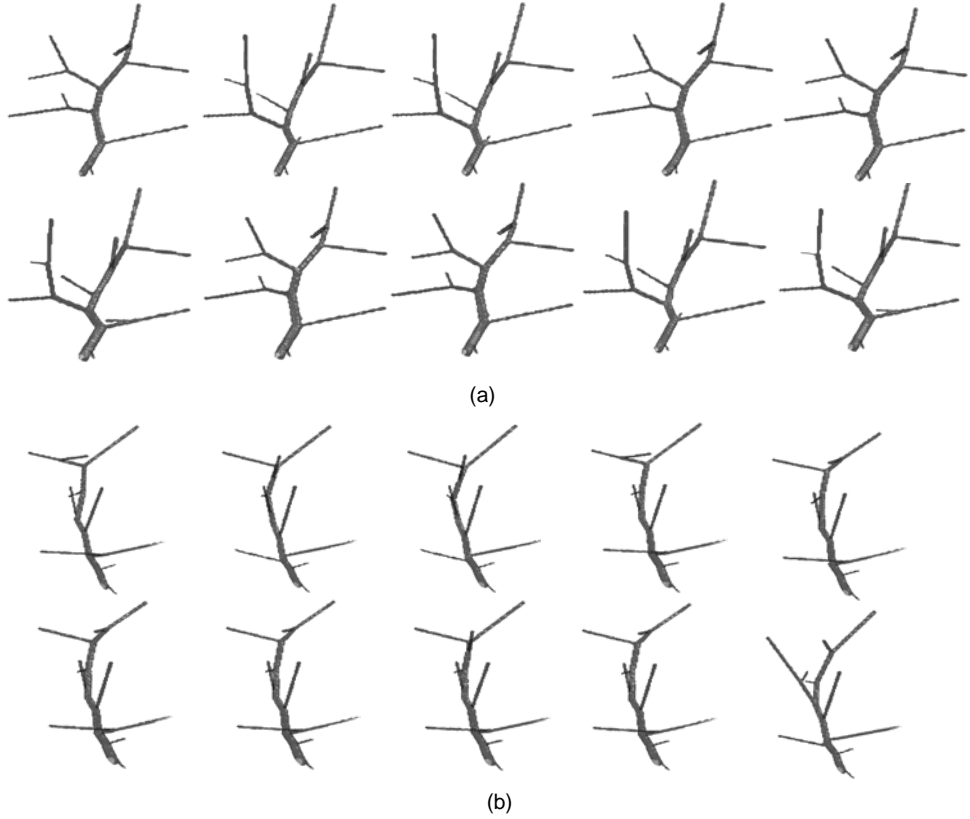

(a)

(b)

Fig. 6. Trees of dataset 1 from group 1. (a) Set 1, (b) Set 2.

TABLE 2
PERFORMANCE OF TAMBAC ON 2-ARY TREES VERSUS 3-ARY TREES

| Data group | Tree order | Tree depth | Branching pattern | Geometric attribute | TAMBAC with NCut accuracy | TAMBAC with K-means accuracy |
|---|---|---|---|---|---|---|
| 1 | 2 vs 3 | 3 vs 3 | Different | U[2,5] vs U[4,7] | 1.00±0.02 | 0.99±0.04 |
| | 2 vs 3 | 3 vs 3 | Different | U[2,5] vs U[7,10] | 1.00±0.02 | 1.00±0.02 |
| | 2 vs 3 | 3 vs 3 | Different | U[2,5] vs U[10,15] | 1.00±0.01 | 1.00±0.01 |
| 2 | 2 vs 3 | 3 vs 3 | Random | U[2,5] vs U[4,7] | 0.73±0.09 | 0.71±0.10 |
| | 2 vs 3 | 3 vs 3 | Random | U[2,5] vs U[7,10] | 0.75±0.18 | 0.70±0.09 |
| | 2 vs 3 | 3 vs 3 | Random | U[2,5] vs U[10,15] | 0.74±0.10 | 0.70±0.09 |
| 3 | 2 vs 3 | 3 vs 5 | Different | U[2,5] vs U[4,7] | 1.00±0.00 | 1.00±0.00 |
| | 2 vs 3 | 3 vs 5 | Different | U[2,5] vs U[7,10] | 1.00±0.00 | 1.00±0.00 |
| | 2 vs 3 | 3 vs 5 | Different | U[2,5] vs U[10,15] | 1.00±0.00 | 1.00±0.01 |
| 4 | 2 vs 3 | 3 vs 5 | Random | U[2,5] vs U[4,7] | 0.66±0.07 | 0.65±0.07 |
| | 2 vs 3 | 3 vs 5 | Random | U[2,5] vs U[7,10] | 0.67±0.07 | 0.66±0.09 |
| | 2 vs 3 | 3 vs 5 | Random | U[2,5] vs U[10,15] | 0.67±0.07 | 0.67±0.08 |

The results of clustering 2-ary versus 3-ary trees with variations in both topology and geometric attributes are presents in Table 2. When the branching pattern is "different", the difference in tree order leads to a further improvement of the accuracy for both NCut and K-means. However, when the branching pattern is "random", the simulated tree topology is more complicated so it is not surprising that the performance of TABMAC with either NCut or K-means decreases. For example, for Groups 1 and 3, TAMBAC achieves an average accuracy of 1, while the average accuracy for Groups 2 and 4 drops to 74% and 67% for TAMBAC with NCut, and 70% and 66% for TAMBAC with K-means.

### 4.2 Performance on Simulated 3D Vascular Trees

Normal blood vessels and tumor-induced vessels are quite different in morphology [10], which motivate us to verify whether TAMBAC can learn the difference in vascular trees. In this section, a publicly available 3D vascular tree synthesis tool called VascuSynth [52] is employed to generate the tree data for our experiments. VascuSynth can generate vascular trees with different topology and geometry according to user-specified angiogenesis factors and fluid dynamics. To be more specific, the topology of a tree is controlled by the so called oxygen demand map (ODM) in VascuSynth, which specifies the oxygen requirement of different spatial locations in tissue (e.g., brain); the tree geometry is regulated by the perforation pressure, flow constraint and volume consistency. There are three parameters and one matrix that we can specify to generate different vascular trees, including the number of branching nodes, $\lambda$, $\nu$, and the ODM map matrix [52]. The number of branching nodes can be modified to change the tree topology. Increasing $\lambda$ will reduce the radii of the vascular branches and increasing $\nu$ will reduce the segment length of the vessels. Changing the ODM map matrix will influence the growing direction and branching position, and thus accordingly change the topology and geometry of the vascular trees.

To evaluate the performance of the TAMBAC method, six groups of 3D vascular trees are generated using different settings of the three control parameters and ODM. Within each group, there are 100 datasets, and each dataset contains 20 vascular trees that belong to two different sets (similar design to that in the previous section). In total, 12,000 vascular trees are generated. Within each dataset, the two sets correspond to two different ODMs, and the parameters of $\lambda$ and $\nu$ also follow different distribution (see Table 3). For illustration purpose, some examples of the generated vascular trees are shown in Fig. 6, which belong to the first dataset from group 1. One can visually tell that the vascular trees from the same set are much more similar to each other than those from the other set. The clustering results are summarized in Table 3, which shows that an average accuracy of 77% and 74% can be achieved over the 12,000 trees using the NCut and

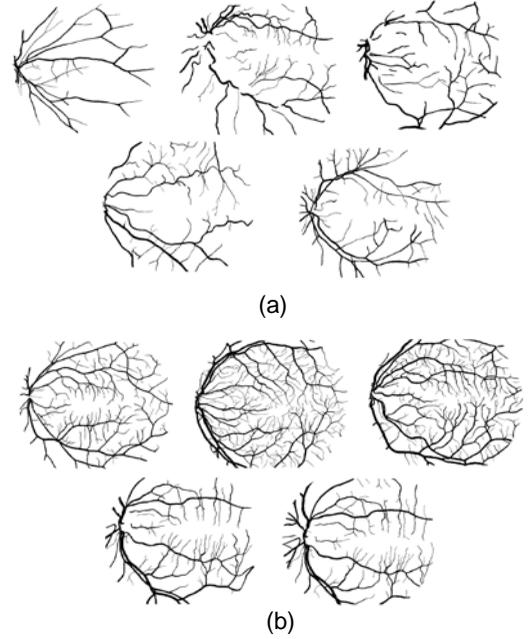

Fig. 7. Ten vessel trees extracted from the retinal images in STARE. (a) Five samples from subjects with retinopathy, (b) Five samples from normal subjects.

K-means methods, respectively. Also, with the increase of the difference in the distributions of $\lambda$ and $\nu$, an increase in the clustering accuracy is observed in Table 3.

TABLE 3
PERFORMANCE OF TAMBAC ON 3D SYNTHESIZED VASCULAR TREES

| Data group | Branching node | ODM | $\lambda$ | $\nu$ | TAMBAC with NCut accuracy | TAMBAC with K-means accuracy |
|---|---|---|---|---|---|---|
| 1 | 10 vs 10 | Different | U[2,3] vs U[2,3] | U[1,2] vs U[1,2] | 0.75±0.13 | 0.71±0.12 |
| 2 | 10 vs 10 | Different | U[2,3] vs U[2,4] | U[1,2] vs U[1,3] | 0.80±0.17 | 0.76±0.11 |
| 3 | 10 vs 10 | Different | U[2,3] vs U[4,5] | U[1,2] vs U[3,4] | 0.79±0.12 | 0.77±0.14 |
| 4 | 10 vs 15 | Different | U[2,3] vs U[2,3] | U[1,2] vs U[1,2] | 0.72±0.11 | 0.71±0.13 |
| 5 | 10 vs 15 | Different | U[2,3] vs U[2,4] | U[1,2] vs U[1,3] | 0.77±0.11 | 0.72±0.11 |
| 6 | 10 vs 15 | Different | U[2,3] vs U[4,5] | U[1,2] vs U[3,4] | 0.79±0.13 | 0.78±0.14 |

## 4.3 Performance on Real Tree Data

To evaluate the performance of the proposed framework on real tree data, we select ten retinal vascular images from the widely-used STARE database [53]. One can tell that the selected images of retinopathy patients (Fig. 7(a)) are visually different from those of normal subjects (Fig. 7 (b)), especially in terms of branching pattern and tree depth. To be specific, the images in Fig. 7(a) are samples 2, 3, 4, 44 and 291 and those in Fig. 7(b) are samples 77, 82, 235, 236 and 255 in STARE.

Considering the performance of TAMBAC with NCut is better than TAMBAC with K-means as suggested in Sections 4.1 and 4.2, here we use TAMBAC with NCut to cluster real trees. Also, to evaluate the influence of the SCNMF step on TAMBAC's performance (recall that like any other NMF method, SCNMF will not always produce the same basis vectors for different runs), five rounds of experiments (each with 100 repetitions) have been conducted for two different numbers of attributes. For each round of experiment, the average classification accuracy and standard deviation are presented in Table 4. If only consider two geometric attributes (vessel segment length and tortuosity), the average clustering accuracy is approximately 73%; if consider three geometric attributes (radius, length, and tortuosity), the average accuracy is around 71%. These two numbers are close to each other, and an explanation for this observation is that the radii of the vessel segments from the two groups are very similar (see Fig. 7). Overall, we can tell from Table 4 that even if SCNMF may produce different basis vectors at different runs, the performance of TAMBAC is consistent and stable.

TABLE 4

PERFORMANCE OF TAMBAC ON STARE DATA

| TAMBAC and two geometric attributes | | | | |
|---|---|---|---|---|
| Accuracy± SD | 0.73±0.11 | 0.72±0.11 | 0.74±0.09 | 0.71±0.12 | 0.73±0.11 |

| TAMBAC and three geometric attributes | | | | |
|---|---|---|---|---|
| Accuracy± SD | 0.71±0.09 | 0.70±0.10 | 0.71±0.11 | 0.71±0.09 | 0.70±0.10 |

TABLE 5

METRIC COMPARISON ON 3D VASCULAR TREES

| Data goup | TAMBAC $d_{L1}$ | TAMBAC ED | QED | Torsello's metric |
|---|---|---|---|---|
| 1 | 0.75±0.13 | 0.73±0.12 | **0.76±0.14** | 0.69±0.15 |
| 2 | **0.80±0.17** | 0.76±0.16 | 0.79±0.18 | 0.71±0.17 |
| 3 | **0.79±0.12** | 0.76±0.12 | 0.78±0.15 | 0.73±0.15 |
| 4 | **0.72±0.11** | 0.68±0.12 | 0.71±0.14 | 0.65±0.13 |
| 5 | **0.77±0.11** | 0.72±0.11 | 0.72±0.13 | 0.67±0.15 |
| 6 | **0.79±0.13** | 0.76±0.14 | 0.77±0.15 | 0.72±0.14 |

TABLE 6

METRIC COMPARISON ON STARE DATA

| Dataset | TAMBAC $d_{L1}$ | TAMBAC ED | QED | Torsello's metric |
|---|---|---|---|---|
| STARE | **0.73** | 0.68 | 0.70 | 0.60 |

## 4.4 Comparison with Other Representative Metrics

To further verify the performance of TAMBAC, two state-of-the-art tree distance metrics are compared, including the newly-proposed QED [22] and Torsello's polynomial-time metric (denoted by $d_4$) [50]. The reason of choosing these two metrics is that Feragen et al. [22] have shown that QED has many desirable theoretical properties (e.g., existence and uniqueness) in comparison with other metrics like tree editing distance (TED), and Torsello's polynomial-time metric outperforms all the other metrics developed in [50]. The QED and $d_4$ metrics are defined in the original tree space as follows

$$d_{QED}(x, y) = \inf\{\sum_{i=1}^{k} d(x_i, y_i) \big| x_1 \in x, y_i \sim x_{i+1}, y_k \in y\} \,,$$

$$\text{with } d(x_i, y_i) = \|x_i - y_i\|_2 \,,$$

and

$$d_4(x, y) = 1 - \frac{W(\phi_{xy})}{|x| + |y| - W(\phi_{xy})} \,,$$

where $x$ and $y$ are two trees, $\sim$ denotes an equivalence relation, $|x|$ is the cardinality of the node set of tree $x$, and $W(\phi_{xy})$ is the overall similarity induced by the maxi-mum similarity subtree isomorphism computed by Hungarian algorithm. For fairness of comparison, all the steps in the TAMBAC pipeline remain the same, except that at Step 6 in Algorithm 1, the distance metric used in NCut can be $d_{L1}$, QED or Torsello's metric; also，Step 5 is not needed for QED and Torsello's polynomial-time metric since they are defined and calculated only in the original tree space. In addition, the Euclidean distance (denoted as TAMBAC ED in the result tables) is also directly used in the meta-tree space to provide more evidence of using $d_{L1}$ as a better approximation to the geodesic.

The comparison is performed on both the synthesized 3D vascular trees and the real data form STARE, and the results of clustering accuracy are listed Tables 5 and 6, respectively. From both tables, we can tell that the $d_{L1}$ metric outperforms Euclidean distance, QED and Torsello's metric for almost all the cases.

## 4.5 Algorithm Sensitivity

In reality, two types of noise may occur to tree data, including attribute noise (disturbance in attribute value) and topology noise (structure perturbation of tree). To evaluate the sensitivity of the TAMBAC framework to different noise sources, we perform further experiments on simulated trees. More specifically, two groups of 2-ary trees (group 2 and group 6 from Table 1) are used as the example ground truth data. Three scenarios are considered for this sensitivity evaluation: adding attribute noise only, adding topology noise only, and adding both attribute and topology noise. Considering that the largest tree of depth 3 has at most 15 edges, attribute noise is added to 5, 10 or 15 randomly-selected edges in each tree. The attribute noise follows a truncated Gaussian distribution with a mean zero and a standard deviation of being 30% of the original attribute value. The notations AN(5), AN(10) and AN(15) in Table 7 means attribute noise is added to 5, 10 or 15 edges, respectively. To add topology noise, 1, 3 and 5 additional edges are randomly added to the ground truth trees at a probability of 0.5 and with their geometric attribute values randomly sampled from 2 to 5, which are similar to the attribute values of the ground truth trees. The three cases of different topology noise (structure perturbation) are denoted as TN(1), TN(3) and TN(5) in Table 7. The combination of AN(5) & TN(1), AN(10) & TN(3), and AN(15) & TN(5) have also been considered to evaluate our method. TAMBAC with NCut is used for all the cases in Table 7. It can be clearly seen from Table 7 that the clustering accuracy of our framework only decrease marginally as the noise level increases (the largest decease in clustering accuracy is about 10%). These results suggest that our method is not very sensitive to both attribute and topology noise.

TABLE 7
CLUSTERING ACCURACY OF TAMBAC UNDER DIFFERENT NOISE CONDITIONS

| Data group | Clustering accuracy | Clustering accuracy | Clustering accuracy | Clustering accuracy |
|---|---|---|---|---|
| **Attribute noise** | **Clean data** | **AN(5)** | **AN(10)** | **AN (15)** |
| **2** | 1.00±0.01 | 0.99±0.05 | 0.98±0.05 | 0.99±0.06 |
| | 1.00±0.02 | 0.99±0.02 | 0.99±0.02 | 0.98±0.07 |
| | 1.00±0.01 | 0.98±0.06 | 0.98±0.08 | 0.99±0.04 |
| **6** | 0.78±0.11 | 0.78±0.10 | 0.76±0.11 | 0.77±0.11 |
| | 0.93±0.11 | 0.87±0.14 | 0.85±0.15 | 0.90±0.13 |
| | 0.98±0.07 | 0.92±0.10 | 0.92±0.09 | 0.93±0.11 |
| **Topology noise** | **Clean data** | **TN(1)** | **TN(3)** | **TN(5)** |
| **2** | 1.00±0.01 | 0.98±0.07 | 0.93±0.12 | 0.89±0.15 |
| | 1.00±0.02 | 0.99±0.06 | 0.98±0.07 | 0.99±0.03 |
| | 1.00±0.01 | 1.00±0.01 | 1.00±0.01 | 1.00±0.02 |
| **6** | 0.78±0.11 | 0.74±0.10 | 0.74±0.12 | 0.73±0.12 |
| | 0.93±0.11 | 0.92±0.11 | 0.90±0.12 | 0.93±0.12 |
| | 0.98±0.07 | 0.95±0.10 | 0.93±0.11 | 0.96±0.09 |
| **Attribute & topology noise** | **Clean data** | **AN(5) & TN(1)** | **AN(10) & TN(3)** | **AN(15) & TN(5)** |
| **2** | 1.00±0.01 | 0.98±0.08 | 0.93±0.12 | 0.90±0.13 |
| | 1.00±0.02 | 0.98±0.06 | 0.99±0.04 | 0.99±0.05 |
| | 1.00±0.01 | 0.98±0.05 | 0.98±0.05 | 0.99±0.03 |
| **6** | 0.78±0.11 | 0.78±0.10 | 0.73±0.13 | 0.69±0.13 |
| | 0.93±0.11 | 0.89±0.13 | 0.87±0.15 | 0.89±0.13 |
| | 0.98±0.07 | 0.93±0.10 | 0.93±0.09 | 0.92±0.10 |

## 5 CONCLUSIONS

A framework called TAMBAC is developed in this study for tree-structured data clustering based on a novel tree data parameterization (called TAMP) and the structure-constrained nonnegative matrix factorization (SCNMF). TAMP can represent both the topological and geometric information of trees in forms of matrices such that the tree-structured data analysis problem can be solved in the matrix manifold realm. This representation and the physical structure constraints in tree data naturally lead to the development of SCNMF. Based on the factorization re-sults, a meta-tree space is constructed and within this space, we explore the distance metrics and incorporate both NCut and a Fréchet-mean-based K-means method to accomplish the tree clustering task. Our simulation studies and real data experiments have clearly shown the efficacy and accuracy of the proposed framework.

We also recognize the limitations of this framework. First, the support tree in the TAMBAC framework is constructed from a given population of trees, which may need to be regenerated to accommodate changes in tree topology when merging new data. Second, similar to all other component decomposition based methods like PCA and ICA, the meta-tree bases in TAMBAC are determined from existing data, which may need to be recomputed when new data are incorporated into existing data. However, the two problems mentioned above are addressable (e.g., by introducing a sufficiently large support tree or incorporating incremental learning techniques), and we will continue to improve and extend TAMBAC in our future work.

## Acknowledgments

This work is supported in part by NIH grant HHSN272201000055C. Non-US support included the National Natural Foundation of China grant 61105034, Research Fund for the Doctoral Program of Higher Education of China grant 20100201120040, China Postdoctoral Science Foundation grant 20110491662, 2012T50805.

**Na Lu** received her B.S. and Ph.D. degrees from Xi'an Jiaotong University, Xi'an, Shaanxi, China in 2002 and 2008, respectively. Currently, she is an Assistant Professor at Xi'an Jiaotong University, Xi'an, Shaanxi, China. Her research interests include statistical image analysis, machine learning, cognitive science and robotics.

**Hongyu Miao** received his M.S. and Ph.D. degrees in Mechanical Engineering from University of Rochester, Rochester, NY in 2004 and 2007, respectively. He is currently Associate Professor of the Department of Biostatistics, University of Texas Health Science Center at Houston, Houston, TX. His research interests include mathematical modeling, statistical image analysis, computational biology and biological applications.